\documentclass[final,1p,times]{elsarticle}

\usepackage{amssymb}

\usepackage{graphicx}
\usepackage{amsmath}
\usepackage{amssymb}
\usepackage{caption}
\usepackage{subcaption}
\usepackage[table]{xcolor}
\usepackage{multirow}
\usepackage{multicol}
\usepackage{colortbl} 
\usepackage{xcolor}
\usepackage{booktabs}
\usepackage{xcolor}
\usepackage{tcolorbox}

\usepackage{lineno}

\journal{Pattern Recognition}

\begin{document}

\begin{frontmatter}



\title{\textcolor{black}{ReBaR: Reference-Based Reasoning for Robust Pose Estimation from Monocular Images}}


\author[a,c]{Yongkang Cheng }
\author[b]{Mingjiang Liang \corref{cor}}
\author[a]{Jifeng Ning }
\author[d]{Gaoge Han }
\author[b]{Wei Liu }
\author[c] {Shaoli Huang }

\cortext[cor]{Corresponding author Mingjiang Liang. E-mail address:  breeliang@gmail.com.}



\affiliation[a]{organization={College of Information Engineering Northwest A\&F University},
            city={Yangling},
            postcode={712100}, 
            state={Shaanxi},
            country={China}}

\affiliation[b]{organization={University of Technology Sydney},
            city={Sydney},
            postcode={2007}, 
            country={Australia}}

\affiliation[c]{organization={Tencent AI Lab},
            city={Shenzhen},
            postcode={518057}, 
            state={Guangdong},
            country={China}}

\affiliation[d]{organization={City University of Hong Kong},
            city={Hongkong},
            postcode={999077}, 
            country={China}}
\begin{abstract}
This paper introduces a novel method, ReBaR (\textbf{Re}ference-\textbf{Ba}sed \textbf{R}easoning for Robust Human Pose and Shape Estimation), designed to estimate human body shape and pose from single-view images. ReBaR effectively addresses the challenges of occlusions and depth ambiguity by learning reference features for part regression reasoning. Our approach starts by extracting features from both body and part regions using an attention-guided mechanism. Subsequently, these features are used to encode additional part-body dependencies for individual part regression, with part features serving as queries and the body feature as a reference. This reference-based reasoning allows our network to infer the spatial relationships of occluded parts with the body, utilizing visible parts and body reference information. ReBaR outperforms contemporary methods on three benchmark datasets and still maintains competitive advantages among recent new approaches. Demonstrating significant improvement in handling depth ambiguity and occlusion. These results strongly support the effectiveness of our reference-based framework for estimating human body shape and pose from single-view images.
\end{abstract}




\end{frontmatter}



\begin{figure*}[t]
  \begin{center}
  \vspace{-2em}
    \includegraphics[width=1.0\textwidth]{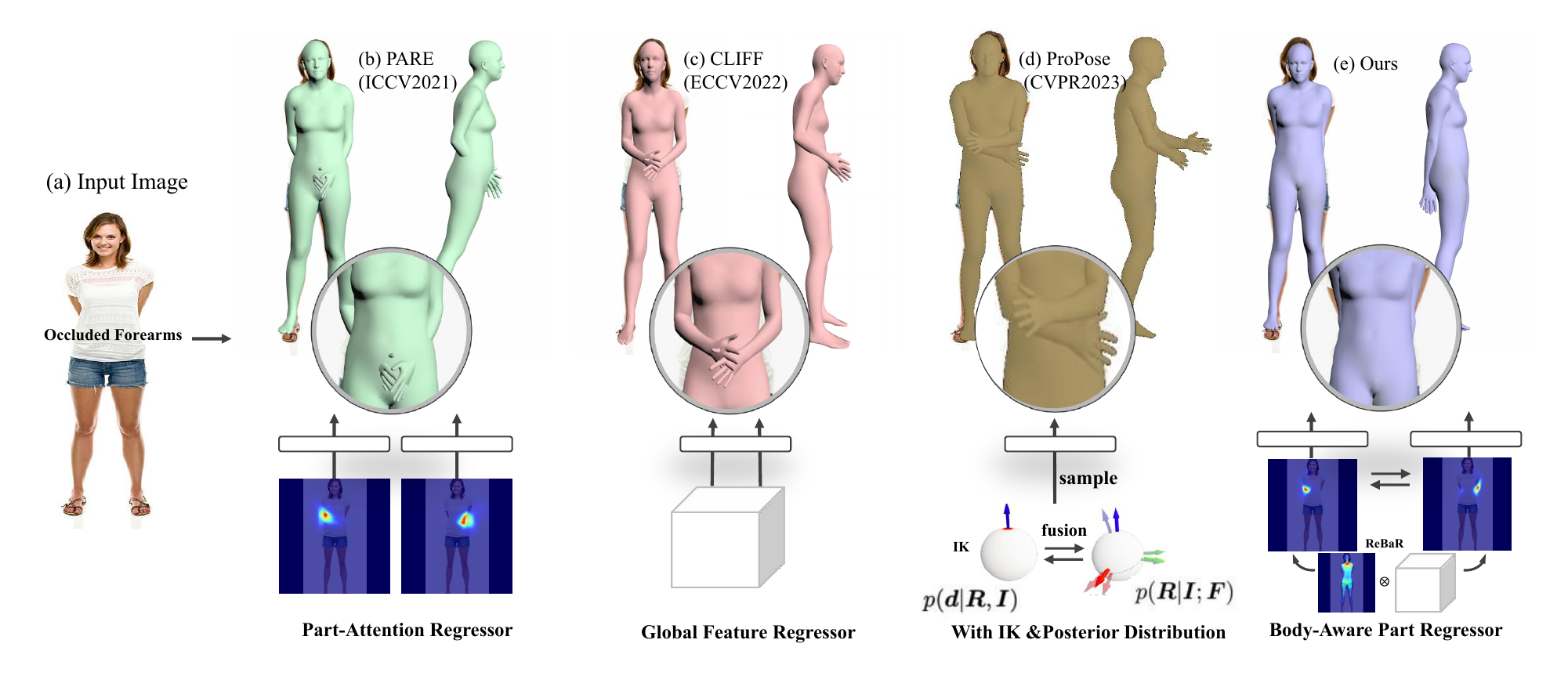}
    \vspace{-1em}
  \end{center}
  \caption{Illustration of our method's main superiority over existing approaches. When given a challenging input (a), PARE (b)~\citep{kocabas2021pare}, CLIFF (c)~\citep{li2022cliff} and ProPose (d)~\citep{fang2023learning} struggle to accurately estimate the pose of the forearm due to occlusion and depth ambiguity. Our method (e), on the other hand, tackles these challenges by exploring reference - based reasoning for part regression.}
  \label{fig:title_img}
  \end{figure*}
  
\section{Introduction}
\label{sec1}

Human body shape and pose estimation from monocular images has become increasingly important in computer vision due to its wide range of applications in fields such as human-computer interaction, virtual reality, and digital human animation. This task involves reconstructing the human body by obtaining parameters of a human body model (such as SMPL~\citep{loper2023smpl}) from single RGB images. Regression-based methods~\citep{kanazawa2018end} that predict human model parameters from image features have made significant progress in recent years and have become the leading paradigm. However, they still face several limitations.

One major limitation is the handling of severe occlusions. Occlusion occurs when a body part is obscured by another object or body part, making it difficult to estimate its pose accurately. Most existing methods based on global feature learning usually struggle with occlusions, which limits their performance in real-world scenarios. Recent work~\citep{kocabas2021pare} has attempted to alleviate this issue by leveraging neighboring visible parts to improve the estimation of occluded parts. However, this strategy is unreliable as it may mistake the pose prediction of an adjacent visible part as the occluded one in some cases, for example, when a leg is impeded mainly by one arm. Another limitation is the handling of depth ambiguity. Depth ambiguity occurs when the relative depth of two body parts cannot be determined from a single image, leading to errors in pose estimation. This is particularly challenging in monocular RGB settings together with self-occlusion. For example, as shown in Figure~\ref{fig:title_img}, given a human with hands behind the back, existing methods fail to estimate a reasonable pose of the forearms. 

However, humans can effortlessly deduce that the hands are behind the back in such an image by utilizing a combination of whole-body visual cues. This suggests that humans rely on a form of reference-based reasoning when perceiving objects in images to handle conclusion and depth ambiguity. Inspired by human ability, we introduce the ReBaR framework. It uses reference-based reasoning to combine global visual cues for better pose estimation. This approach effectively handles occlusions and depth ambiguity, leading to accurate human body shape and pose estimation from single-view images, even in difficult situations.

Our method primarily involves an attention-guided encoder for feature extraction and a body-aware regression module for part representation. The attention-guided encoder, based on ConvNets, focuses on body and part features while considering 2D and 3D dependencies between parts. The body-aware regression module uses a two-layer transformer to encode body-aware part features for each part and a one-layer transformer for per-part regression. This approach allows our method to incorporate both local visual cues and global information from the entire body, improving estimation accuracy by inferring the spatial relationship of occluded parts with the body.

Our experiments first show that our method consistently outperforms state-of-the-art approaches on 3DPW~\citep{von2018recovering} and Human3.6M~\citep{ionescu2013human3} datasets. 
Then, we evaluate our method on 3DPW-OCC~\citep{von2018recovering} and 3DOH~\citep{zhang2020object} datasets and demonstrate its effectiveness in occlusion handling. We also compare axis MJE on the 3DPW-TEST dataset. The result shows that our method's performance gain over state-of-the-art methods is attributed mainly to the much lower error of Z-axis MJE, demonstrating that it effectively reduces depth ambiguity in estimation. Finally, our qualitative results and ablation study confirm the proposed method's effectiveness. \textcolor{brown}{Our precise reconstruction data is highly effective for various generation tasks, such as MMOFusion~\citep{wang2025mmofusion}.}

\textcolor{black}{The main contributions of this paper are as follows:}
\begin{itemize}
  \item  \textcolor{black}{We are inspired by human perception and propose the ReBaR framework. We use reference-based reasoning to combine global visual cues, thereby addressing the issues of occlusion and depth ambiguity and enhancing the accuracy of human pose estimation from single-view images.}
  \item \textcolor{black}{We construct an attention-guided encoder and a body-aware regression module. These modules integrate local visual information and global body information to infer the spatial relationships of occluded parts.}
\end{itemize}


\section{Related Works}
\label{sec2}


\subsection{Regression-based methods}
Estimating human mesh and pose~\citep{liu20243d} from single images~\citep{atrevi2017very} has attracted increasing attention in the computer vision community.  The mainstream methods in this task include optimization, point cloud-based and regression-based. Point cloud-based methods~\citep{krawczyk2023segmentation}, such as HPE~\citep{xu2022head}, directly predict coordinates in 3D space to facilitate human body reconstruction. Optimization-based approaches concentrate on the optimization algorithms to fit parametric models (e.g., SMPL~\citep{loper2023smpl}) based on 2D observations such as keypoints and silhouettes. These methods have been criticized for having long optimization processes and high sensitivity to initialization~\citep{kolotouros2019learning}. Therefore, regression-based approaches  have become a more popular paradigm in recent years, as they can directly and fast regress parameters from image features.  HMR~\citep{kanazawa2018end} is a milestone work that introduces reprojection loss of keypoints as weak supervision, enabling training regression models on in-the-wild datasets.  Recent work continues to explore more advanced weak supervision loss, such as providing pseudo 3D ground truth~\citep{joo2020exemplar}, real-time regression based on multi-scale dense connections~\citep{luvizon2009ssp} and multi-Person Pose Estimation~\citep{zhao2020cluster}. For instance, SPIN~\citep{kolotouros2019learning} incorporates the optimization process into the regression framework, allowing better-optimized 3D results as supervision for the network. MHFormer++~\citep{li2023multi} and MHCanonNet~\citep{kim2024mhcanonnet} employ multi-scale attention mechanisms to address robust pose estimation. CLIFF~\citep{li2022cliff} proposes considering the box information and computing the reprojection loss from the original images, which facilitates predicting global rotations. However, all these methods employ global representation for regression, making them sensitive to occlusion and partially visible humans.  

\subsection{Occlusion handling}
Handling occlusion is challenging yet crucial for human shape and pose estimation.  Simulating data containing occlusion is a straightforward way towards this issue. This line of work attempts to generate training data of occluded human bodies by cropping images~\citep{joo2020exemplar}, overlaying object, and augmenting feature maps. Despite the success achieved in reducing occlusion sensitivity, the realisticness of synthetic data is usually poor, resulting in unsatisfactory performance in dealing with real occlusion data. The other line of work relies on visibility cues to aid in dealing with occlusion. \cite{cheng2019occlusion} obtains the keypoint visibility label and masks the loss of invisible points in training, encouraging the network to infer occluded joints from visible parts. \cite{yao2022learning} proposes VisDB, which first trains a network to predict the coordinates and visibility label of the mesh vertex and then incorporates them to regress the SMPL parameters. However, VisDB is a two-stage framework and requires a test-time optimization procedure to obtain the final results.  In contrast, PARE~\citep{kocabas2021pare} is an end-to-end learning framework that  infers the occluded parts by exploiting the attention mechanism to find helpful information from neighboring visible parts. However, the visual cues of adjacent visible parts sometimes are unreliable or insufficient to infer the occluded parts.  Unlike these approaches, Our method leverages visible parts and body reference information to infer the occluded parts.

\section{Preliminary Work}\label{sec3}
\subsection{SMPL model}
Our approach utilizes the SMPL parametric model to represent the human body. The model requires two essential characteristics: pose, denoted as $\theta \in \mathbb{R}^{72}$, and shape, denoted as $\beta \in \mathbb{R}^{10}$. The SMPL model produces a positioned 3D mesh $\mathcal{M}(\theta,\beta)\in \mathbb{R}^{6890\times3}$ as a differentiable function. The reconstructed 3D joints are obtained as $\mathcal{J}_{3D}=\mathcal{W} \mathcal{M} \in \mathbb{R}^{J\times 3}$, where $J=24$ and $\mathcal{W}$ is the pre-trained linear regression matrix.

\subsection{Part/Body segmentation maps}
\begin{figure*}
  \centering
  \includegraphics[width=0.85\textwidth]{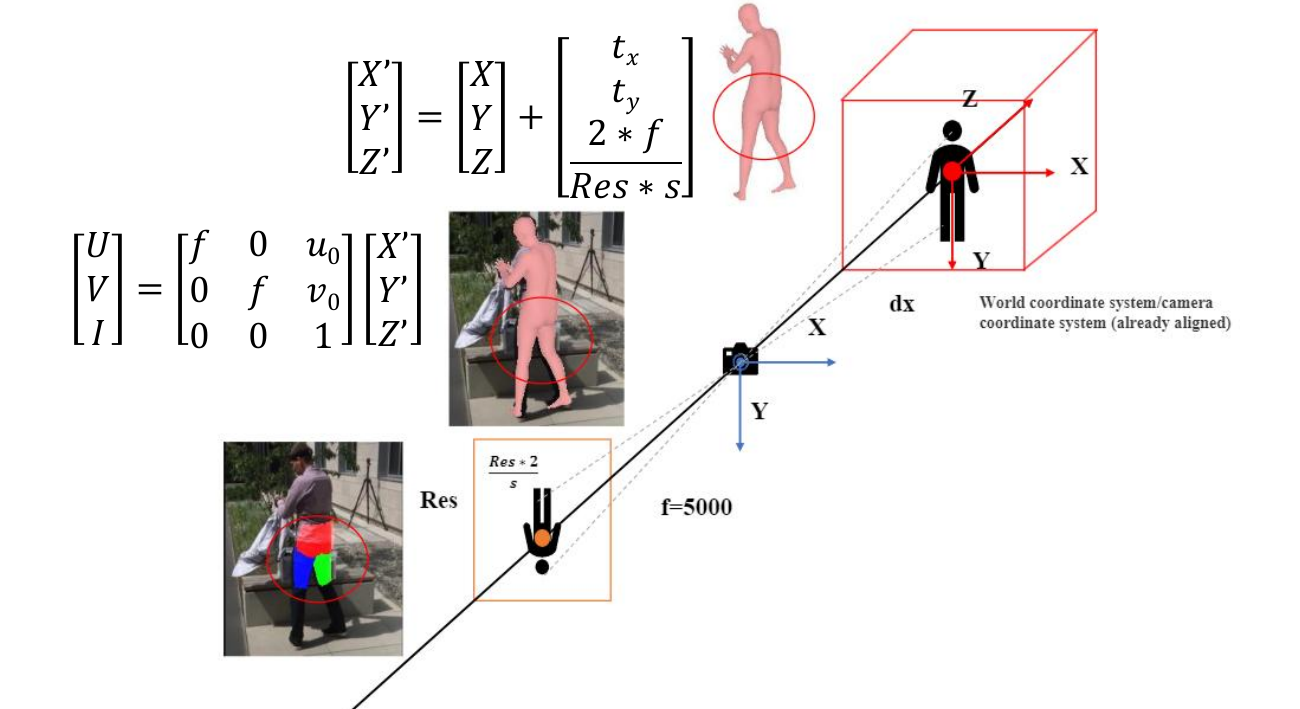}
  \caption{Weak-perspective projection illustration.}
  \label{fig:weakpp}
\end{figure*}

To obtain segmentation maps for the auxiliary supervision of the body attention map and part attention map in the Attention-Guided Encoder (AGE) module, segmentation labels are required for each part of the original image. However, labeling the original image is a time-consuming and costly process. Luckily, we can utilize the SMPL model to obtain the vertex coordinates in the camera coordinate system that correspond to each image. Using weak perspective transformation techniques~\citep{kissos2020beyond}, we can then generate the 2D vertex coordinates in the pixel coordinate system that are needed to generate the segmentation map. This method enables us to obtain the necessary segmentation maps without the need for costly and manual labeling of the original image. \textcolor{black}{Considering that the dataset we used contains data collected by devices and fitted data, the camera parameters in some datasets are incomplete. For this part of the data, we introduce inverse weak-perspective projection to generate body/part attention map labels.} First, let's discuss weak-perspective projection:\\
The weak-perspective projection, as described in ~\cite{kissos2020beyond}, is based on the premise that the focal length and object distance are sufficiently large, allowing for the neglect of variations in the object along the Z-axis. It is a technique for converting three-dimensional camera coordinates into pixel coordinates. Before performing the projection, the 3D keypoints are normalized to a cube in the range [-1, 1], and the camera is aligned to the world coordinate system origin (the cube's center). Next, the projection camera parameters ($s, t_{x}, t_{y}$) are required, where s represents the human body's scale in the cropped image, and $t_{x}$ and $t_{y}$ represent the translation within the cube. In this study, the cropping size Res is 224. To perform the weak-perspective projection, we first increase the object distance, as shown in the following formula:
\begin{equation}
\begin{aligned}
  t_{z}=\frac{2 \times f}{Res \times s},
\end{aligned}
\end{equation}
where, f represents the focal length, which is set to 5000 in our study. Subsequently, we add $t_{x}$, $t_{y}$, and $t_{z}$ as translations to the 3D keypoints and introduce the following formula for weak-perspective projection:
\begin{equation}
\begin{aligned}
\quad
\begin{bmatrix} U\\V\\I \end{bmatrix}=\begin{bmatrix} f & 0 &u_{0} \\0&f&v_{0}\\0&0&1 \end{bmatrix} \begin{bmatrix} X+t_{x}\\Y+t_{y}\\Z+t_{z} \end{bmatrix},
\quad
\end{aligned}
\end{equation}

\noindent where we set $u_{0}$ and $v_{0}$ to the image's center point position, and f represents the focal length. The resulting $U$ and $V$ are the 2D keypoints in the pixel coordinate system after projection. In the reverse process, we obtain 3D mesh points and 3D/2D keypoints using the ground truth (GT) labels and the SMPL model. We then compute the translation in reverse using the 3D/2D keypoints. Subsequently, we add the translation to the 3D mesh points in the same manner and perform weak-perspective projection to obtain body/part attention map labels. The process is shown in Figure~\ref{fig:weakpp}.

\begin{figure*}[h]%
  \begin{center}
    \includegraphics[width=1.0\textwidth]{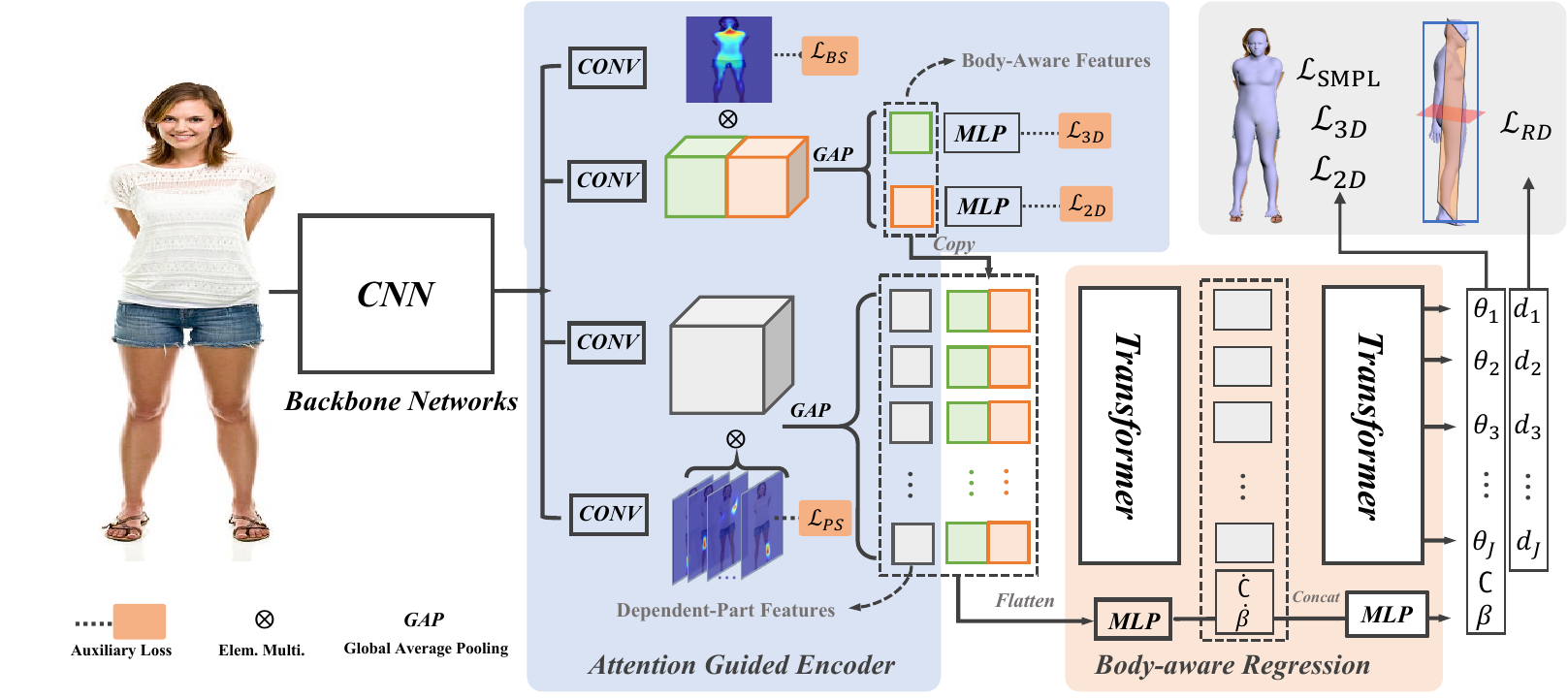}
  \end{center}
  \caption{\textcolor{black}{ReBaR network architecture. Given an input image, our method first extracts body and part features based on a soft attention mechanism. Then each part feature is concatenated with the body feature as an input token to the transformer to encode body-aware part features for camera prediction and SMPL parameter regression. Among them, the green feature block $F_{2D}$ is used to regress the 2D key point coordinates, the orange feature block $F_{3D}$ is used to regress the 3D key point coordinates. The large three-dimensional gray cube in the blue AGE module represents the 3D features of the entire body. This cube is then element-wise multiplied with the body part attention maps to generate the 3D features for each body part, as shown by the small gray feature blocks at the output of the blue module. In BAR, global and local features are fused to output globally aware features, as shown by the gray features in the orange module.}}
  \label{fig:arc}
  \end{figure*}

\section{Approach}\label{sec4}

As depicted in Figure \ref{fig:arc}, our method mainly includes an attention-guided encoder and a body-aware regression module. The first component aims to learn proper attention maps to locate informative regions for extracting parts and body features. The second module encodes body-aware part features based on these output features and feeds them into a one-layer transformer for camera and SMPL parameter regression. Subsequently, it generates the outputs of camera parameters, SMPL parameters, \textcolor{black}{and the explicit relative depth constraints we proposed}. We introduce our method in more detail as follows. 

\begin{figure*}
  \centering
  \includegraphics[width=0.75\textwidth]{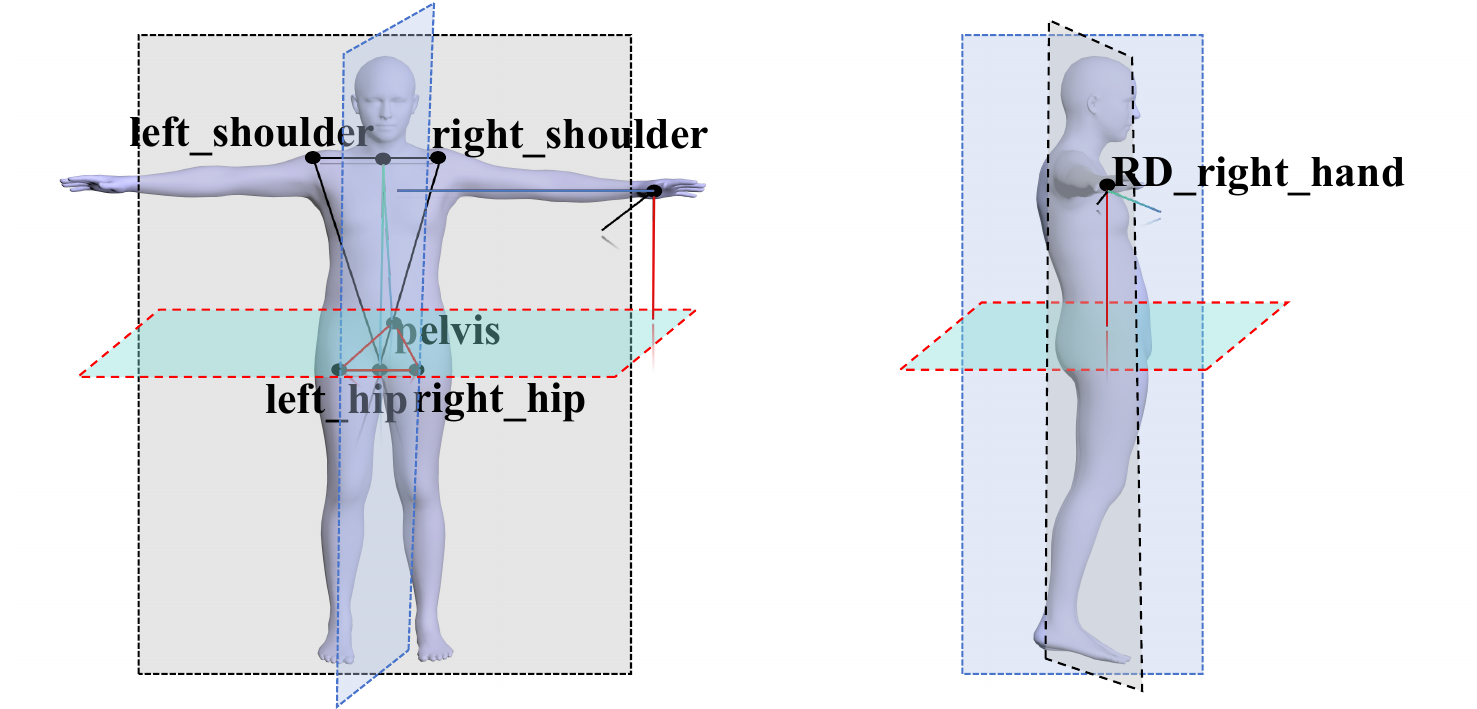}
  \caption{Illustration of the calculation of the relative depth from the torso plane. Shows the torso points for constructing the torso planes and the relative depth of the right hand.}
  \label{fig:rp}
\end{figure*}
\textcolor{black}{
\subsection{Torso plane and Relative depth}
In human pose and shape regression, previous methods have relied on optimizing axis-angle representations based on 3D keypoint supervision, indirectly constraining the pose space. The advantage of this indirect constraint is its ability to capture global pose information, thus maintaining overall pose consistency. However, it may fail to accurately capture local joint depth information, especially in cases of occlusion or depth ambiguity between joints. To address this issue, we simultaneously construct a 3D plane coordinate system for the torso to compute the relative depth of each joint. By directly supervising joint depth information, we can more accurately recover local joint spatial relationships, thereby improving the accuracy of pose and shape regression. This direct constraint compensates for the inadequacy of indirect constraints in capturing local joint depth information. Additionally, the relative depth loss encourages the model to learn depth patterns more applicable to various poses and body shapes. As it focuses on the relative relationship between joints and planes, the model is less likely to overfit specific training examples and can better adapt to unseen data, enhancing its generalization capability. As shown in figure~\ref{fig:rp}, we select the shoulder, hip, and pelvis points on the human torso to establish the frontal, lateral, and transverse planes of the torso. The center points of the skin regions of each joint are defined as the actual joint positions, and joint axes are defined by bringing equations greater than 0 into the plane as positive directions, and the depth value is calculated based on the distance from the point to the plane. This is used as a basic fact label for relative depth, and explicit plane consistency constraints are constructed.
}

\subsection{Attention-guided encoder}
The human body is a complex structure with various parts and joints that have different contributions to the overall shape and pose. To accurately model and reconstruct the 3D human body from a single RGB image, it is crucial to focus on the informative regions and capture both global and local features. The Attention-Guided Encoder (AGE) is designed to achieve this by learning proper attention maps and extracting features from different body parts and the whole body. We use a CNN backbone to extract feature volumes from the input image. From there, we leverage four convolutional layers to obtain the body attention map, body feature volume, part attention map, and part feature volume, respectively.

\noindent \textbf{\textcolor{black}{Reference Feature.}} 
\textcolor{black}{To establish the dependency relationship between the 2D space \textcolor{black}{(green block)} and 3D space \textcolor{black}{(orange block)} of the reference features, we concatenate the reference feature blocks of the two spaces to construct a global reference feature block $F_{\text{body}}\in \mathbb{R}^{H\times W\times (C+C')}$. Then we extract the attention-guided reference features $F_{\text{brf}}\in \mathbb{R}^{(C+C')\times 1}$ by performing the Hadamard operation on the body attention map and the global reference feature block, and introduce global 2D/3D keypoint information for supervision:}
\begin{equation}
\begin{aligned}
  {F_{\text{brf}}}=\sigma({{Att}_{\text{body}}})^{T} \odot {{F_{\text{body}}}},\\
\end{aligned}
\end{equation}
where $\odot$ is the Hadamard product and $\sigma(Att_{\text{body}})$ is used as a soft attention mask to aggregate features. By using this approach, we can better capture the global features of the human body and establish the relationship between 2D and 3D.

\noindent \textbf{Dependant-part Features.}
Another task of the AGE module is to extract part-query features. To accomplish this, we feed the feature volume extracted by the backbone network to two convolutional layers, which respectively extract part-3D features \textcolor{black}{(gray block, which is consistent with the part feature blocks in PARE.)} $F_{\text{part}}\in \mathbb{R}^{H\times W\times C''}$ and part-attention maps $Att_{\text{part}}\in \mathbb{R}^{H\times W\times 24}$ (where $C''$ is the feature dimensions of the part-3D feature block). Follow the Pare method~\cite{kocabas2021pare}, we use part-segmentation maps as an auxiliary supervision to aid the learning process. Similar to the extraction process for Attention-guided reference feature, we perform a Hadamard operation on the part-feature block and the part-attention map to extract part-query features$F_{\text{pqf}}\in \mathbb{R}^{C''\times 24}$. This approach allows us to focus on the relevant parts of the body and extract features that are important for accurate 3D reconstruction.

\subsection{Body-aware Regression}
The relationships between different body parts play a vital role in understanding the human body's 3D structure and pose. A simple concatenation of part features may not be sufficient to capture these relationships and dependencies. The Body-Aware Regression module (BAR) is introduced to encode body-aware part features for each part while establishing associations between parts. This module helps the model to leverage the spatial relationships between body parts and the whole body, leading to a more accurate 3D reconstruction. These features are then utilized as input to the regression transformer, which predicts SMPL model parameters, camera parameters, and relative torso plane depth. Finally, the model parameters in the output are fed into the SMPL model to generate a 3D human body.

To encode body-aware part features for each part, we concatenate the Attention-guided reference feature from the AGE module to the part-query features and establish a set of tokens $T=\{t^{1}_{bap},t^{2}_{bap},t^{3}_{bap},...,t^{23}_{bap},t^{24}_{bap} \}$(where $t^{i}_{bap}=F^{i}_{pqf}\oplus F_{brf},\oplus$ means concatenate operation.) as the input of the two-layer body-aware transformer. Using this encoder, we query part features from body feature to encode body-aware part features $F_{\text{bapf}}\in \mathbb{R}^{24\times 128}$, building part representations and dependencies to reference body. The resulting body-aware part features sequence can not only capture the unique characteristics of each part but also combine relevant global information from the entire body, which is crucial for accurate 3D reconstruction. By using the transformer to establish dependencies between the parts and reference body, our feature encoder can effectively exploit the spatial relationship between parts and bodies to generate more informative part representations. Overall, our feature encoder is capable of finding useful information from parts and the whole body and outputs a set of body-aware part features that are highly informative for the subsequent regression task. Specifically, the formula is as follows:
\begin{equation}
\begin{aligned}
  Q &=W^{q}T;K=W^{k}T;V=W^{v}T, \\
  F_{\text{bapf}} &=LN(T+LN(softmax(\frac{QK^{T}}{\sqrt{d_{k}}})V)),
\end{aligned}
\end{equation}
where $W$ is the weight, Q, K and V are the matrices after linear mapping of input, and $dk$ is the sequence length which is used to transform the attention matrix into a standard normal distribution.

To help regression transformers better learn and perceive the correlation between parts and bodies, we propose a reference plane consistency loss. We select the shoulders, hips and pelvis points on the human torso to establish the frontal plane of the torso, the side plane of the torso and the cross-section of the torso respectively (detailed diagram can be found in supplementary materials). In order to correspond to the 24 parts in the AGE module, we input the label parameters into the SMPL model to generate $gt\_vertices$ $\in \mathbb{R}^{6890\times 3}$, and take the center of each part area as the joint point. Define the joint point to bring into the plane equation greater than 0 as the positive direction, and calculate the depth value according to the distance from the point to the plane. Use this as the ground-truth label for relative depth, and construct explicit planar consistency constraints. This constraint helps the regression transformer to learn the spatial information between parts and bodies to solve the problem of depth ambiguity. Next, we take the output of the body-awareness encoder as the input to a single-layer regression transformer to learn dependencies among the features of each body-aware part. The final BAR module outputs a set of SMPL model parameters, camera parameters, and relative depth $RD\in \mathbb{R}^{24\times 3}$ to the torso plane. We use the standard loss and the reference plane consistency loss for each part for training the network.

\subsection{Loss Functions}
We train ReBaR as a supervised learning problem, aiming to minimize the discrepancy between predicted outputs and ground truth annotations. To achieve this, we define a set of loss functions that capture different aspects of the model's performance. Given a training dataset $D$ which containing $N$ images, we have various ground truth data, including RGB images $I \in \mathbb{R}^{w \times h \times 3}$, SMPL parameters $(\hat{\theta}\in \mathbb{R}^{24 \times 3},\hat{\beta} \in \mathbb{R}^{10})$, relative depths $\hat{RD} \in \mathbb{R}^{24\times 3}$, 3D and 2D coordinates $(\hat{J_{3D}} \in \mathbb{R}^{J \times 4, \hat{J_{2D}} \in \mathbb{R}^{J \times 3})}$ of body joints $J$, and segmentation labels for whole bodies and body parts $Seg_{\text{parts}} \in \mathbb{R}^{H \times W \times 24}$ and $Seg_{\text{body}} \in \mathbb{R}^{H \times W \times 1}$.

To supervise body and parts attention, we project the vertices generated by the SMPL model onto pixel coordinates, creating segmentation labels ${Seg_{\text{body}}} \in \mathbb{R}^{H \times W \times 1}$ and ${Seg_{\text{parts}}} \in \mathbb{R}^{H \times W \times 24}$. This allows us to supervise attention at the pixel level, providing more detailed information for evaluating model performance.\\
We use pose parameters $\theta$ and shape parameters $\beta$ to output body vertices $V_{3d} \in \mathbb{R}^{6890 \times 3}$. To compute the keypoint loss, we need the SMPL 3D joints $J_{3D}(\theta,\beta)=WV_{3d}$, which are computed using a pretrained linear regressor $W$. With the inferred weak-perspective camera, we compute the 2D projection of the 3D joints $J_{3D}$, as $J_{2D} \in \mathbb{R}^{J \times 2}=s\Pi(RJ_{3D})+t$,where $R\in SO(3)$ is the camera rotation matrix and $\Pi$ is the orthographic projection. We use the combination of loss functions to train ReBaR, including AGE loss and BAR loss, where each term is calculated as:
\begin{equation}
\begin{aligned}
\mathcal{L}_{\text{AGE}} &= \mathcal{L}_{\text{Aux2D}}+\mathcal{L}_{\text{Aux3D}}+(\mathcal{L}_{\text{BS}}+\mathcal{L}_{\text{PS}}),\\
\mathcal{L}_{\text{BAR}} &= \mathcal{L}_{\text{2D}}+\mathcal{L}_{\text{3D}}+\mathcal{L}_{\text{SMPL}}+\mathcal{L}_{\text{RD}},\\
\mathcal{L}_{\text{joints}} &= ||J_{\text{joints}}-\hat{J_{\text{joints}}}||_{F}^{2},\\
\mathcal{L}_{\text{SMPL}} &= ||\Theta-\hat{\Theta}||_{2}^{2},\\
\mathcal{L}_{\text{RD}} &=||\text{RD}-\hat{\text{RD}}||_{2}^{2},\\
\mathcal{L}_{\text{BS/PS}} &= \frac{1}{\text{HW}}\sum\limits_{h,w}\limits^{}CrossEntropy(\sigma(\text{Att}_{h,w}),{\text{Seg}_{h,w}}),\\
\end{aligned}
\end{equation}
\textcolor{black}{Among them, the parameter symbol with a hat represents the ground-truth label of the corresponding parameter, such as $J_{joint}$ is the prediction of our model and $\hat{J_{joint}}$ is the ground-truth label of 3D keypoints.} $\mathcal{L}_{\text{joints}}$ is the supervision of all $3D$ and $2D$ keypoints. In addition, auxiliary supervision losses are $\mathcal{L}_{\text{Aux2D}}$, $\mathcal{L}_{\text{Aux3D}}$, $\mathcal{L}_{\text{BS}}$ and $\mathcal{L}_{\text{PS}}$. Both $\mathcal{L}_{\text{Aux2D}}$ and $\mathcal{L}_{\text{Aux3D}}$ are calculated using the formula $\mathcal{L}_{\text{joints}}$, while $\mathcal{L}_{\text{BS}}$ and $\mathcal{L}_{\text{PS}}$ represent attention map constraints. We train the model with all losses for the first 100 epochs. After that, we remove the segmentation supervision loss and continue training the model. This initial training phase helps the model learn to recognize and emphasize relevant regions in the input images. Once the attention mechanism has been guided towards the body parts, we remove the segmentation supervision loss and continue training. During this phase, the attention mechanism adapts further to better recover the human mesh from the images.

\section{Experiments}\label{sec5}

\subsection{\textcolor{black}{Implementation and Datasets}}
In our network, we adopt the HRNet32/48~\cite{cheng2020higherhrnet} network trained on MPII~\cite{andriluka20142d} for estimating 2D keypoints as our backbone, which exhibits faster convergence speed compared to the pre-trained models based on ImageNet. For the regressor part guided by the attention mechanism based on the Transformer Encoder in the body-aware regression module, we adopt a 12-layer Transformer Encoder for the part-global perception guide, and in the regressor, we introduce a 6-layer self-attention encoder to regress both the SMPL rotation parameters and the relative trunk depth information. Our network adopts an input size of $224 \times 224$, and we apply standard data augmentation techniques during training, such as random rotation, scaling, horizontal flipping, and cropping. We train the network using the widely adopted Adam optimizer, with a batch size of 256 and a learning rate of $5 \times e^{-5}$ . Specifically, we train our model with a batch size of 64 on a single V100 GPU, where all losses are jointly optimized and all constraints are supervised. Then we fine-tune the model for 30 epochs with a mixture of Human3.6M~\cite{ionescu2013human3}, in-the-wild data, and 3DHP~\cite{kissos2020beyond} at a ratio of (0.6, 0.3, 0.1). We follow the training strategy of PARE, supervising the ReBaR module only during the first 10 epochs, then releasing the explicit constraint on the attention map, which is subsequently used for evaluating the occluded 3DPW-OCC data. Finally, we fine-tune the model for an additional 10 epochs on 3DPW training data and apply the best model for evaluating 3DOH~\cite{zhang2020object}.\\
\noindent\textbf{Datasets.}
\noindent We train our model on five publicly available datasets, including \textcolor{black}{COCO(74K)~\cite{joo2020exemplar}}, MPII (14K)~\cite{andriluka20142d}, MPI-INF-3DHP (90K)~\cite{kissos2020beyond}, Human3.6M (292K)~\cite{ionescu2013human3}, and 3DPW (68K)~\cite{von2018recovering}. Human3.6M is an indoor dataset characterized by simplistic scenes, precise labels, and ease of learning. \textcolor{black}{By contrast, COCO is an outdoor dataset with complex scenes, lower image quality and a lack of SMPL labels. Thanks to the hint from CLIFF ~\cite{li2022cliff}, we obtained pseudo SMPL labels through the fitting algorithm. However, in order to maintain a consistent comparison with PARE, we also presented the training results on MS COCO which only has 2D keypoint labels. Subsequently, we used the method of EFT~\cite{joo2020exemplar} to provide SMPL pseudo-labels for the COCO dataset. The specific process was consistent with that of CLIFF}, which further enriched the labels of the COCO dataset. These supplementary annotations offer a more comprehensive understanding of the underlying human pose and shape, which is beneficial for human reconstruction. The weight distribution of our supervision functions is as follows: segmentation map supervision, reference feature 2D/3D auxiliary constraints, and relative depth supervision all have a weight of 60, while the SMPL parameter and keypoint supervision are five times greater than the former. This balanced approach contributes to the model's effectiveness.  In the process of training ReBaR, we respectively use global 2D/3D keypoints, body/part attention map and relative depth supervision to assist the network to learn body-aware part features. For the supervision of body/part attention, we only supervise on the  dataset, and reset the weight of $L_{\text{BS/PS}}$ in the loss function to 0 on the mix and 3DPW datasets. For the AGORA dataset, we do not supervise the relative depth.

In the evaluation phase, we compared the effectiveness of our method on four public datasets. We conducted a comparison with all monocular reconstruction methods of the same period on 3DPW, Human3.6M, and AGORA respectively. For the AGORA dataset, which is designed for complex environments and multi-erson scenarios, we first pre-trained the network on COCO (74K), and then continuously fine-tuned it on a mixed and 3DPW dataset. Subsequently, we further fine-tuned the model on AGORA~\citep{patel2021agora} and evaluated its performance on the AGORA test set.



\begin{table}[ht!]
	\centering
\begin{tabular}{lcccccc}
        \toprule[1.5pt]
		\multirow{2}{*}{Method} & \multicolumn{3}{c}{3DPW-ALL}   & \multicolumn{3}{c}{3DPW-TEST} \\
		& MJE$\downarrow$ & PAMJE$\downarrow$ & MVE$\downarrow$ & MJE$\downarrow$ & PAMJE$\downarrow$ & MVE$\downarrow$  \\ \hline
		I2L-MeshNet~\citep{moon2020i2l} & 93.2 & 58.6 & - & -& -&- \\
            METRO~\citep{lin2021end} &- &- &- &77.1 &47.9 &88.2\\
            HybrIK~\citep{li2021hybrik} & 80.0 & 48.8 & 94.5 & 74.1 & 45.0 & 86.5\\
            FastMETRO–L–H64~\citep{cho2022cross} &- &- &- &73.5 &44.6 &84.1\\
            \textcolor{black}{NIKI+~\citep{li2023niki}} & \textcolor{black}{-} & \textcolor{black}{-}  & \textcolor{black}{-} & \textcolor{black}{71.3} & \textcolor{black}{40.6} & \textcolor{black}{86.6}   \\
            \hline
		HMR~\citep{joo2020exemplar} &85.1 & 52.2 & 118.5 &- &- &-\\
		ROMP~\citep{sun2021monocular} & 82.7 & 60.5 & - & 76.7 & 47.3 & 93.4\\
		PARE~\citep{kocabas2021pare} & 82.0 & 50.9 & 97.9 & 74.5 & 46.5 & 88.6\\
        \textcolor{black}{OSX~\citep{lin2023one}} & \textcolor{black}{-} & \textcolor{black}{-}  & \textcolor{black}{-} & \textcolor{black}{74.7} & \textcolor{black}{45.1} & \textcolor{black}{-} \\
            VisDB~\citep{yao2022learning} &- &- &- & 72.1 &44.1 &83.5\\
            CLIFF(HR-W32)~\citep{li2022cliff} & \textcolor{black}{83.9} & \textcolor{black}{49.7} & \textcolor{black}{92.5} & \textcolor{black}{73.1} & \textcolor{black}{45.5} & \textcolor{black}{83.8}\\ 
            CLIFF*(HR-W48)~\citep{li2022cliff} & - & - & - & 69.0 & 43.0 & 81.2\\ 
            MotionBERT* (HR-W48)~\citep{zhu2022motionbert} & - & - & - & 68.8 & 40.6 & 79.4\\
            ProPose (HR-W32)~\citep{fang2023learning} & \textcolor{black}{80.9} &\textcolor{black}{47.0} & \textcolor{black}{89.7} & \textcolor{black}{71.8} & \textcolor{black}{41.9}& \textcolor{black}{82.2}\\
            ProPose* (HR-W48)~\citep{fang2023learning} & - & - & - & 68.3 & 40.6 & 79.4\\
            TokenHMR~\citep{dwivedi2024tokenhmr} &- &- &- &\textcolor{black}{71.0} &\textcolor{black}{44.3} &\textcolor{black}{84.6}\\
            MH-HMR~\citep{xuan2024mh} &- &- &- &\textcolor{black}{83.7} &\textcolor{black}{50.5} &- \\
            Multi-HMR(ViT-L)~\citep{baradel2024multi} &- &- &- &\textcolor{black}{\textbf{64.6}} &\textcolor{black}{43.8} &\textcolor{black}{79.7}\\
            \hline
            \textcolor{black}{\textbf{ReBaR (ViT-L)}} &- &- &- &\textcolor{black}{65.1} &\textcolor{black}{\textbf{39.8}} &\textcolor{black}{\textbf{77.5}} \\
            \textcolor{black}{\textbf{ReBaR (HR-W32)}} & \textcolor{black}{80.7} & \textcolor{black}{47.0} & \textcolor{black}{92.3} & \textcolor{black}{71.7} & \textcolor{black}{42.6} & \textcolor{black}{83.4} \\
		\textbf{ReBaR* (HR-W32)} & 76.1 & 45.9 & 90.9 & 69.1 & 41.8 & 81.9  \\
		\textbf{ReBaR* (HR-W48)} & \textbf{72.0} & \textbf{45.3} & \textbf{86.5} & 67.2 & 40.8 & 78.7 \\ 
        \bottomrule[1.5pt]
	\end{tabular}
 \caption{\textcolor{black}{Performance comparison on the 3DPW dataset. For the evaluation on 3DPW - ALL, all methods were trained without using the 3DPW dataset. Bold indicates the best results. "*" indicates the use of the COCO - EFT dataset, and "+" indicates the method that uses its own designed dataset.}}
	\label{tab:3dpw}
\end{table}

\subsection{Comparison to the state-of-the-art}

We first comprehensively compare ReBaR with two types of state-of-the-art methods, including model-free and model-based methods on the 3DPW dataset. In this experiment, we report two evaluation results, namely 3DPW-ALL and 3DPW-TEST. Here,  all methods do not include the 3DPW training dataset in training to evaluate the 3DPW-ALL setting, while for the 3DPW-TEST, they all fine-tune the network on the 3DPW training dataset. In addition, we provide the results of two network structures (HR-W48 and HR-W32) on these two settings, respectively. Table~\ref{tab:3dpw} presents the comparison results with various methods. Our method's performance using the HR-W32 backbone network has surpassed almost all existing methods in all metrics but is only slightly lower MVE than the CLIFF method using the HR-W48 backbone network. However, when using the HR-W48 backbone network, our method significantly outperforms the previous state-of-the-art method CLIFF on 3DPW-TEST, and MJE is reduced from 69.0 to 67.2. It is also worth mentioning that the evaluation results in 3DPW-ALL can better reflect the generalization ability of the method. Yet, our method also outperforms the previous state-of-the-art method HybrIK by a large margin (MJE from 80.0 to 72.0) in this setting.

Among the newer methods, CLIFF, MotionBert, and Propose did not use the original COCO dataset in their original papers. Instead, they used the EFT strategy to generate pseudo-labels for training. The poor 3D human body annotations in the original COCO dataset limited the model's reconstruction accuracy, while the addition of EFT-fine-tuned pseudo-labels improved performance from the original 71.7 to 69.1. Specifically, EFT is a pseudo-label generation method combining regression models and optimization algorithms, which fits 2D key points through the SMPL model to optimize original model weights. In our experiments, we used the SMPL parameters initialized by HMR as input, employed the real annotated 2D key point labels as supervision information, and optimized the weights of the original HMR through the optimization formula:
\begin{equation}
\begin{aligned}
    \hat{\omega}=\underset{\omega}L_{2D}(\pi(M(\sigma_{\omega}(I))),\hat{J_{2D}}) + \lambda(L_{shape}(\beta)),
\end{aligned}
\end{equation}
where $l_{2D}$ is the 2D reprojection loss and $L_{shape}$ is the shape prior loss. EFT introduces a pose prior encoder $\sigma_{\omega}$ during the optimization process to map image features into a reasonable 3D pose space. This ensures that the optimized pseudo-pose labels better comply with human kinematic constraints during fine-tuning, avoiding the generation of unreasonable joint angles.

To ensure a fair comparison, we retrained CLIFF and ProPose using the original COCO labels and included the results in the table for comparison. It can be observed that without using EFT pseudo-labels, CLIFF's performance drops significantly while ProPose's declines slightly. Our method still outperforms both under the original labels.

It is worth noting that the original Multi-HMR uses a pure Transformer structure, with its backbone being ViT-L14. To ensure fairness, we replaced the original HRNet backbone with ViT-L14 to maintain consistency with Multi-HMR. As shown in the table~\ref{tab:3dpw}, our method outperforms Multi-HMR on ViT-L14 in both PAMJE and MVE metrics.

We also evaluate our method on Human3.6M and multi-person dataset AGORA and compare it with state-of-the-art methods. Furthermore, since camera parameter estimation and kids model play a crucial role in the performance of the AGORA dataset, we compare our method with those without kids model and using weak-perspective projection in training for a fair comparison. Table~\ref{tab:agora} shows the result where our method still performs better than multi-person-based approaches and previous state-of-the-art methods.

\begin{table}[ht!]
\centering
\resizebox{0.85\linewidth}{!}{\begin{tabular}{llccccccc}
\toprule[1.5pt]
\multirow{2}{*}{Training} & \multirow{2}{*}{Method} & \multicolumn{2}{c}{AGORA} & \multicolumn{2}{c}{Human3.6M} \\ 
& & MJE$\downarrow$  & V2V$\downarrow$  & MJE$\downarrow$  & PAMJE$\downarrow$\\ \hline
&SPIN~\citep{kolotouros2019learning} &175.1 &168.7 &- &-\\
&PyMAF~\citep{zhang2021pymaf}&174.2 &168.2 &- &- \\
Unfinetuned&EFT~\citep{joo2021exemplar} &165.4 & 159.0 &- &- \\
&PARE~\citep{kocabas2021pare} & 146.2 & 140.9 & - & -\\ 
&\textbf{ReBaR (Ours)} & \textbf{134.6} & \textbf{128.9} & - & - \\ \hline
&ROMP~\citep{sun2021monocular} & 108.1 & 103.4 & - & -\\
&BEV~\citep{sun2022putting} & 105.3 & 100.7 & - & - \\
&FastMETRO–L–H64~\citep{cho2022cross} &- &- &52.2 &33.7\\
Finetuned&VisDB~\citep{yao2022learning} &- &- &51.0 & 34.5\\
&Hand4Whole~\citep{moon2022accurate} & 89.8 & 84.8 & - & -\\ 
&CLIFF~\citep{li2022cliff} & 81.0 & 76.0 & 47.1 & 32.7\\ 

&\textbf{ReBaR} & \textbf{79.9} & \textbf{74.5} & \textbf{45.4} & \textbf{32.1} \\ \bottomrule[1.5pt]
\end{tabular}}
\caption{Performance comparison on the AGORA and Human3.6m dataset. In the upper segment, we present results from a model without fine-tuning on the AGORA dataset. In the lower portion, we showcase the refined outcomes after the fine-tuning process.}
\label{tab:agora}
\end{table}

\subsection{Improved occlusion handling and Reduced depth ambiguity}

\begin{table}[ht!]

\centering
\begin{tabular}{lccccccccc}
\toprule[1.5pt]
\multirow{2}{*}{Method} & \multicolumn{3}{c}{3DPW-OCC} & \multicolumn{2}{c}{3DOH} & \multicolumn{3}{c}{3DPW-TEST}\\ 
 & MJE$\downarrow$ & PAMJE$\downarrow$ &MVE$\downarrow$ & MJE$\downarrow$ & PAMJE$\downarrow$ &MJE$_{x}$ & MJE$_{y}$ & MJE$_{z}$\\ \hline
PARE (R50) &91.3 &\textcolor{black}{57.0} &\textcolor{black}{104.8} &\textcolor{black}{65.7} &\textcolor{black}{45.3} &30.3 &28.9 &58.2 \\ 
CLIFF (R50) &\textcolor{black}{88.0} & \textcolor{black}{52.2} & \textcolor{black}{93.5} &\textcolor{black}{63.7} &\textcolor{black}{42.2} &28.8 &27.1 &51.8 \\\hline
\textcolor{black}{ReBaR (R50)} &\textcolor{black}{\textbf{85.2}} &\textcolor{black}{\textbf{50.3}} &\textcolor{black}{\textbf{92.3}} &\textcolor{black}{\textbf{61.5}} &\textcolor{black}{\textbf{41.9}}  &\textbf{28.7} &\textbf{26.3} &\textbf{49.8} \\
\bottomrule[1.5pt]
\end{tabular}
\caption{\textcolor{black}{Evaluation results on the occluded dataset. The training strategies of all methods were kept consistent with that of PARE, and the training was conducted on COCO, Human3.6M and 3DOH. On the right side are the evaluation results comparing CLIFF with PARE on the three axes. It can be seen that the main improvement lies in the significant enhancement along the depth axis (Z-axis). Since the original paper didn't provide the models, all the models were reproduced by ourselves.}}
\label{tab:ddd}
\end{table}


Table \ref{tab:ddd} presents the results of our proposed ReBaR method, the baseline PARE method and the CLIFF method on the occluded datasets 3DPW-OCC and 3DOH. \textcolor{black}{All methods were evaluated on 3DPW-OCC without using the 3DPW training dataset, and all training strategies were kept consistent with those of PARE. Specifically, when compared with PARE and CLIFF, we adopted the original ResNet50 as the backbone network. Since the PARE repository did not provide the training models for the occluded datasets, we reproduced them, and all network parts were kept consistent with those of PARE to maintain a fair comparison. ReBaR outperforms PARE in both MJE (Mean Joint Error) and PAMJE (Pose-Aware Mean Joint Error), and also demonstrates better performance than CLIFF and PARE on 3DOH. Specifically, on the 3DPW-OCC dataset, compared with PARE, ReBaR reduces MJE and PAMJE by 6.6\% and 11.7\%, respectively. On 3DOH, ReBaR improves by 3.4\% compared with CLIFF and by 6.3\% compared with PARE.} 


\begin{figure*}
  \centering
  \includegraphics[width=1\linewidth]{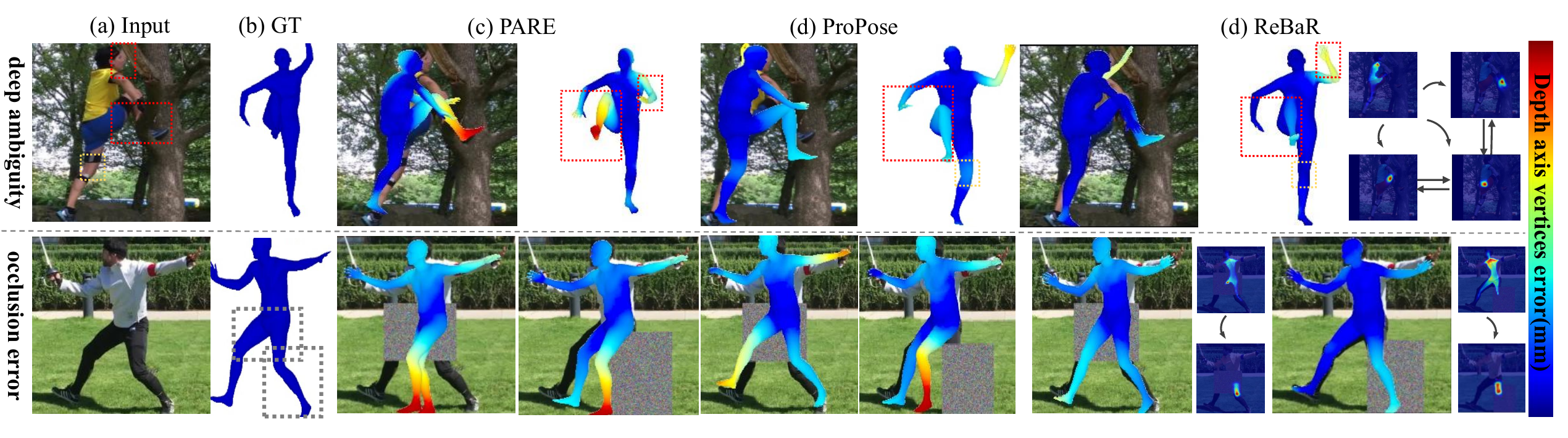}
  \caption{Occlusion and Depth error processing. Compared to PARE and ProPose, our ReBaR exhibits enhanced prediction performance for occluded body parts. From a side view perspective, ReBaR is noticeably superior to PARE in terms of accuracy along the depth axis.}
  \label{fig:occ}
\end{figure*}


To further analyze the depth ambiguity problem in 3D human reconstruction, we evaluated the Mean Joint Position Error (MJE) on the x, y, and z axes on the 3DPW-Test dataset. As shown in Table \ref{tab:ddd}, we can observe that the error on the z-axis is significantly higher than that on the x and y axes in existing methods. This indicates that the depth ambiguity problem in 3D human reconstruction is a major challenge that needs to be addressed. However, our method performs significantly better than the state-of-the-art methods on the z-axis metric, with an improvement of 10\% compared to CLIFF. This result highlights the importance of body-aware part feature encoding, which can infer the spatial relationships between the body and parts through local visual cues around body parts and relevant global information from the whole body, thereby effectively alleviating the depth ambiguity problem.

\textcolor{black}{Inspired by the occlusion sensitivity map proposed by PARE, we took the outdoor data in the 3DPW dataset as an example, calculated the depth error sensitivity map in the same way, and artificially created mosaic occlusions to judge the accuracy of our method in inferring occluded parts from visible parts, as shown in Figure~\ref{fig:occ}. The error value of each mesh point was calculated and marked with colors, where blue represents a lower error compared to the Ground Truth (GT), and red represents a higher error compared to the GT. ReBaR has higher inference accuracy than contemporary methods such as PARE and ProPose, both in the case of occlusion and under depth ambiguity, further demonstrating the effectiveness and robustness of ReBaR.}

\begin{figure}
  \centering
  \includegraphics[width=1\linewidth]{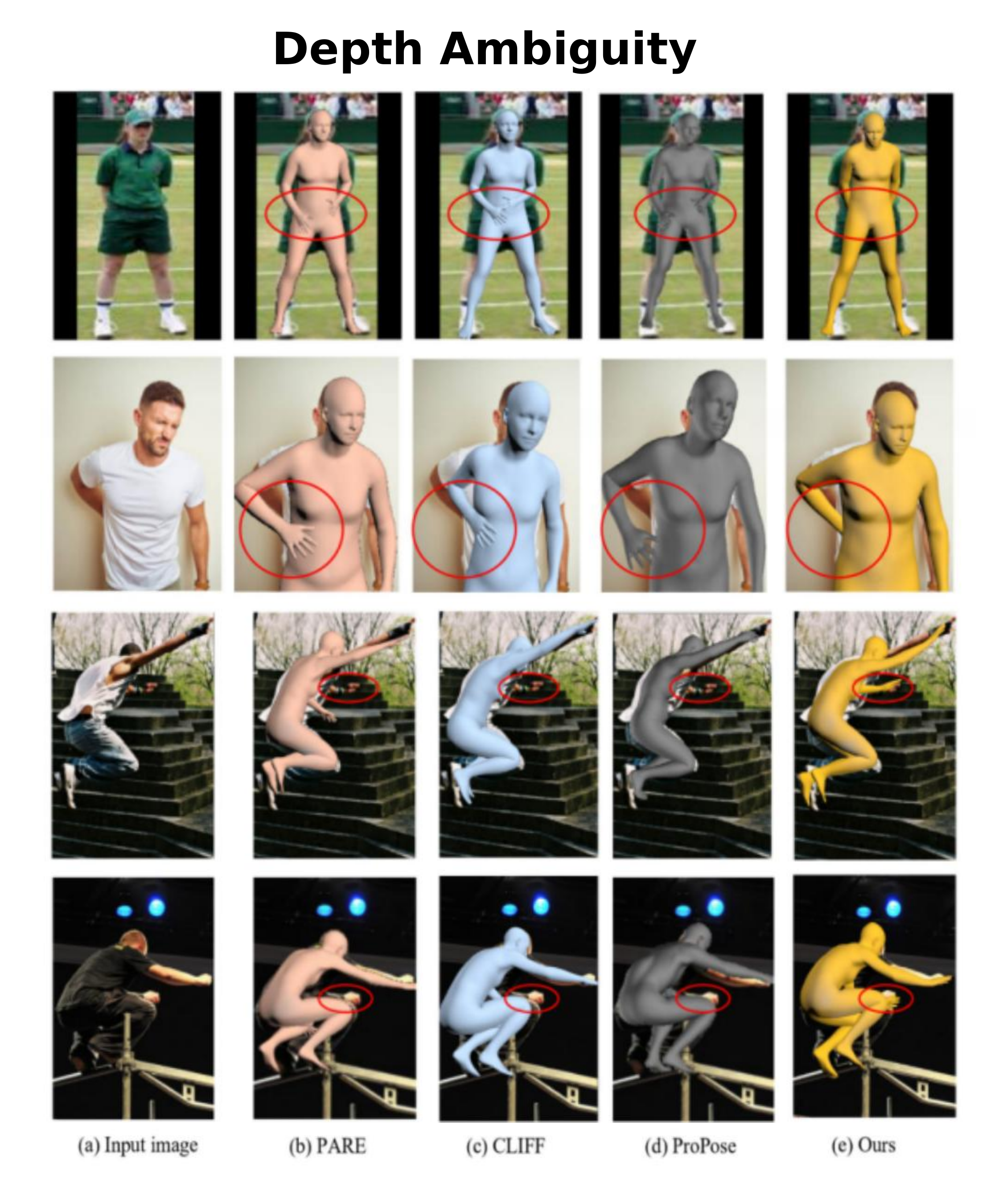}
  \caption{\textcolor{brown}{Qualitative comparison of PARE, CLIFF, ProPose and our method ReBaR in depth-ambiguous and occlusion scenarios.}}
  \label{fig:qua}
\end{figure}

In order to substantiate the superior performance of our method in resolving depth ambiguity and occlusion, we extend our comparisons with the state-of-the-art methods on an in-the-wild dataset. In Figure ~\ref{fig:qua}, we compare the performance of PARE, CLIFF, ProPose and ReBaR on different test datasets. To render the human geometry created by the SMPL model into images, we use the predicted camera parameters and weak perspective projection~\citep{kissos2020beyond}. The results show that PARE, CLIFF and ProPose struggle to accurately infer the depth information between limbs and exhibit ambiguities in the orientation of the human body under large occlusions. However, thanks to the body reference condition dependency established by the BAR module, ReBaR can more accurately recover these motions.

\begin{table}[ht!]

\centering
\resizebox{1.0\linewidth}{!}{\begin{tabular}{lcccc}
\toprule[1.5pt]
{Method } &MJE &PAMJE &MVE & MJE$_{z}$ \\ \hline
Baseline (PARE) &74.5 &46.5 &88.6 &58.2\\
\hline
Baseline+TF &73.2 &44.7 &85.0 &56.6\\
Baseline+HMR-RFeat+MLP &72.7 &43.9 &82.7 &53.1\\
Baseline+HMR-RFeat+TF &71.5 &43.3 &83.8 &49.0\\
Baseline+ATT-RFeat+TF &70.4 &42.4 &82.2 &49.5\\
Baseline+ATT-RFeat+TF+$L_{RD}$ (ReBaR) &69.1 &41.8 &81.9 &46.6 \\
\bottomrule[1.5pt]
\end{tabular}}
\caption{Ablation studies on the primary modules of our method. \textcolor{black}{Here all methods are trained with the same datasets, i.e. COCO, Human3.6M, MPII, MPI-3DHP and 3DPW. }}
\label{tab:abl2}
\end{table}


\subsection{Ablation Study of the Module}


In this experiment, we conducted an ablation study on the main modules of our method. All the models for comparison were using the HR-W32 backbone network. \textcolor{black}{This is because the original PARE paper provided this model, facilitating a fair comparison for us. Consistent with the training strategy of PARE, we trained on the mixed data (COCO, Human3.6M, 3DHP, MPII and 3DPW).} Table \ref{tab:abl2} presents our experimental results. As can be seen from the table, directly combining the PARE method with the Transformer can only bring a slight improvement to PARE (the Mean Joint Error (MJE) is reduced from 74.5 to 73.2). However, after combining with our Attention-guided Reference Features (\textbf{ATT-RFeat}), the overall performance is significantly improved (the MJE is reduced from 74.5 to 70.4, and the MJE Depth is reduced from 58.2 to 47.5). \textcolor{black}{This confirms that our \textbf{ATT-RFeat} effectively solves the problem of depth ambiguity. Meanwhile, the inference strategy based on the reference features and the attention mechanism greatly improves the MJE metric. This is because in the occluded data of 3DPW, our method can infer the invisible parts more robustly and effectively. In addition, the results show that adding the relative distance loss can further improve the overall performance.} We also verified that using the Transformer (\textbf{TF}) to \textcolor{black}{construct the attention relationships among parts for mutual inference performs better than the \textbf{MLP} module in the original PARE. Moreover, although using the HMR features as reference features (\textbf{HMR-RFeat}) can also improve the baseline performance, this further demonstrates the effectiveness of global features as reference features for local inference.}

 Considering the introduction of segmentation map supervision, which is not utilized by some other methods, we added extra ablation experiments to ensure a fair and comprehensive comparison with other techniques. To construct a comparative method that does not use segmentation map supervision (no region attention learning mechanism), we removed all attention-based learning components and expanded the global features extracted from HRNet to match the quantity of SMPL parts. Subsequently, we connected the T-Pose template SMPL parameters as labels to the global features and predicted the SMPL parameters through a self-dimension-reducing Transformer module. We refer to this method as HMR + TF (no add-ons). As shown in Table~\ref{tab:aba2}, this method, which does not rely on attention-based techniques, is at a disadvantage in terms of performance compared to our proposed method. This finding emphasizes the importance of the attention mechanism and the additional supervision provided by ground truth body/part segmentation maps in enhancing performance.

  \begin{table}[ht!]

\centering
\resizebox{0.55 \linewidth}{!}{\begin{tabular}{lcccc}
\toprule[1.5pt]
\multirow{2}{*}{Method} & \multicolumn{3}{c}{3DPW-Test}  \\ 
 & MJE$\downarrow$ & PAMJE$\downarrow$ & V2V$\downarrow$  \\ \hline
HMR (no att.) & 85.1 & 52.2  &118.5\\ 
PARE (+Part-att.) & 74.5 & 46.5  &108.6\\ 
HMR+TF (no att.)& 76.8 & 49.0 &89.9 \\
PARE+TF (+Part-att.) &73.2 &44.7 &85.0\\
ReBaR (+Part-att.) &\textbf{69.1} &\textbf{41.8} &\textbf{81.9}\\ \bottomrule[1.5pt]
\end{tabular}}
\caption{Ablation study of ReBaR on 3DPW-Test. All methods are trained on 3DPW-Train.}
\label{tab:aba2}
\end{table}
 
\subsection{Visual Ablation Experiment on the Reference Feature ATT-RFeat}

\begin{figure}
\centering
  \includegraphics[width=0.55\linewidth]{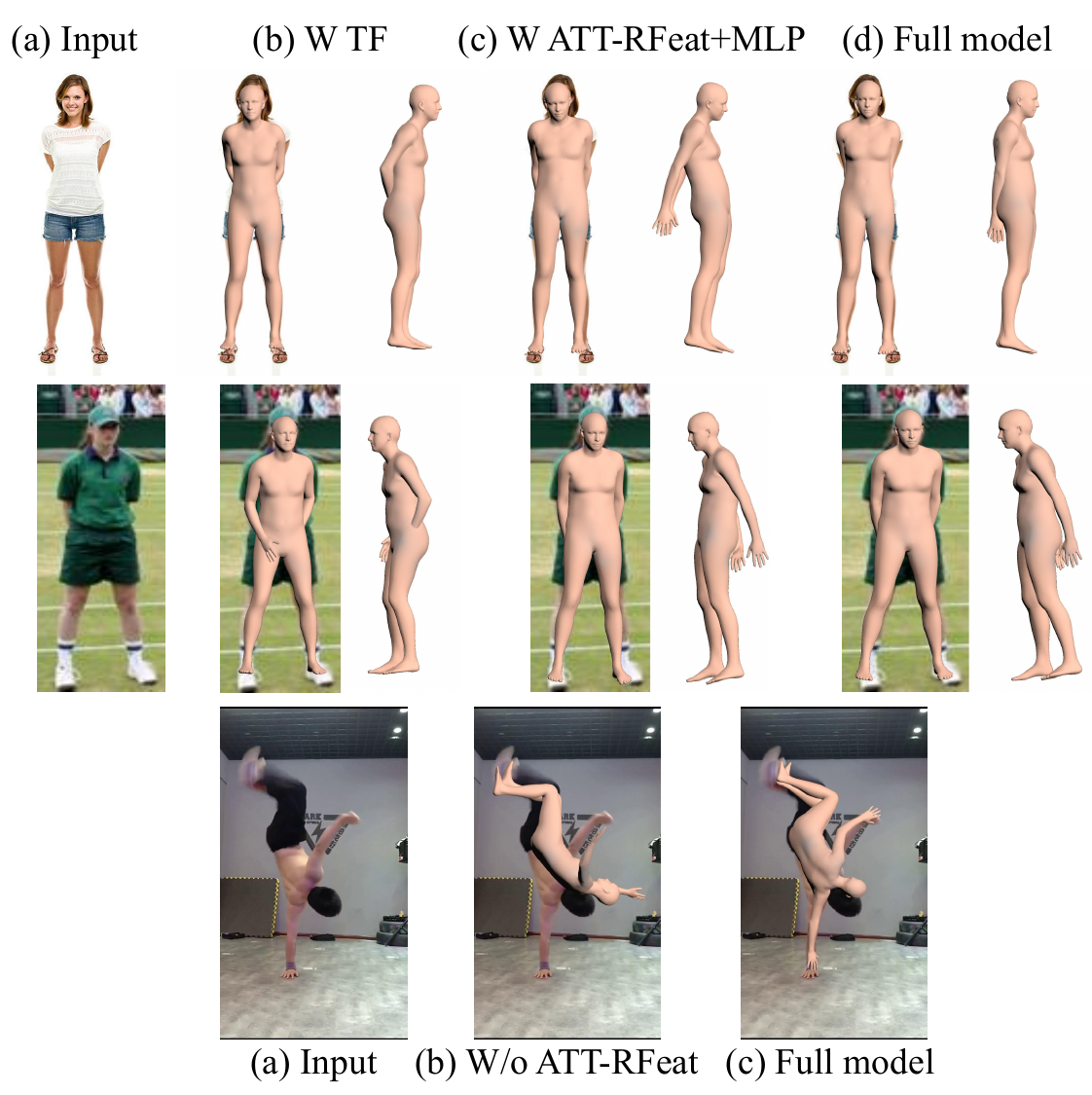}
  \caption{The role of ATT-RFeat. For the first two rows, from left to right: input image, PARE+transformer, PARE+BAR, and the result of the full model. For the last line, from left to right: input image, w/o ATT-Rfeat, and the result of the full model.}
  \label{fig:bar}
\end{figure}

\begin{figure}
\centering
  \includegraphics[width=1.0\linewidth]{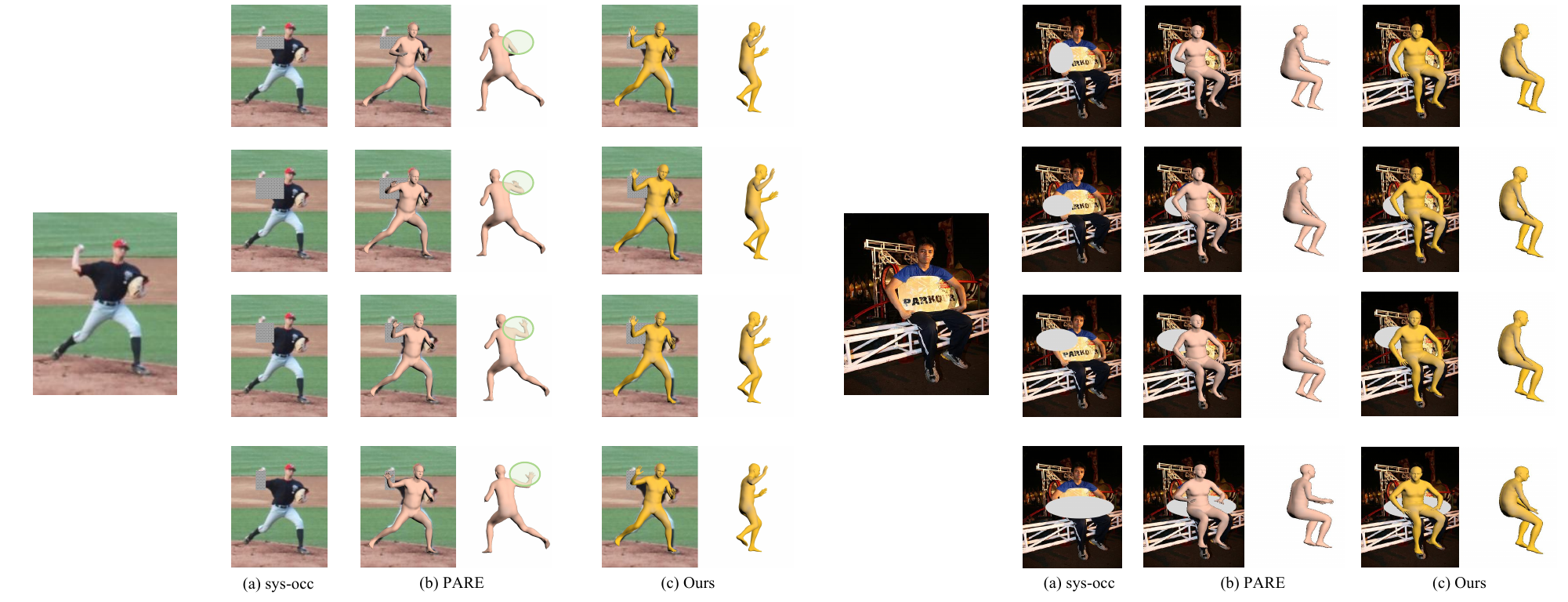}
  \caption{\textcolor{black}{We demonstrate occlusions with different sizes and shapes applied to various body parts, and show the final reconstruction results. As can be seen from the figures, our method exhibits significantly better robustness to occlusions and reconstruction accuracy compared to PARE.}}
  \label{fig:bar}
\end{figure}



\textcolor{black}{ATT-RFeat is the core of our model for file inference, and its role is to utilize the relatively rigid torso features as more reasonable visible information to infer the poses of occluded parts.} This also enables ReBaR to have a better ability to alleviate the depth ambiguity problem compared to existing methods. As shown in Figure \ref{fig:bar}, when we directly associate the component features of PARE, we cannot correctly infer the correct pose of the occluded forearm (for example, the wrong pose where the forearm is bent forward). However, after adding the body reference condition, the model can infer a relatively correct pose, although its depth information is not very accurate. With the help of the relative torso depth constraint, ReBaR can accurately infer the pose of the arm behind.

\section{Conclusion}\label{sec6}
In conclusion, our proposed body-aware part regressor significantly improves human body shape and pose estimation from monocular images compared to current methods. By utilizing a soft attention mechanism to capture global information and dependencies between body parts, and implementing body-aware regression using both local and global information, our method handles severe occlusions and depth ambiguity, achieving state-of-the-art performance across various datasets and scenarios. This work advances human body shape and pose estimation and offers valuable insights for future research. Nonetheless, our method faces challenges with extreme perspectives, causing imprecise torso representations and impacting performance. Future work will explore adaptive reference feature selection.


\section*{Declaration of Competing Interest}
The authors declare that they have no known competing financial interests or personal relationships that could have appeared to influence the work reported in this paper.










\bibliographystyle{elsarticle-num}
\bibliography{egbib.bib}
\end{document}